\documentclass[journal,twoside,web]{ieeecolor}
\usepackage{graphicx}
\usepackage{generic}
\usepackage{cite}

\usepackage{multicol}
\usepackage{amssymb,amsmath, amsfonts}
\usepackage{bm}
\usepackage{algorithm}
\usepackage{physics}
\usepackage{subfigure}
\usepackage{url}
\usepackage{array}
\usepackage{etoolbox}
\usepackage{lcsys}
\usepackage{comment}

\let\classAND\AND
\let\AND\relax
\usepackage{algorithmic}
\let\AND\classAND
\AtBeginEnvironment{algorithmic}{\let\AND\algoAND}

\floatname{algorithm}{Algorithm}

\newenvironment{pf}{\begin{proof}}{\end{proof}}

\newtheorem{thm}{Theorem}
\newtheorem{prob}{Problem}
\newtheorem{remark}{Remark}
\newtheorem{proposition}{Proposition}

\AtEndEnvironment{thm}{\hfill $\square$}
\AtEndEnvironment{prob}{\hfill $\square$}
\AtEndEnvironment{remark}{\hfill $\square$}
\AtEndEnvironment{proposition}{\hfill $\square$}

\newcommand{\calX}{{\mathcal X}}

\newcommand{\Ckk}{C(\xkk)}
\newcommand{\tCkk}{\tilde C(\xkk)}
\newcommand{\eCkk}{c}
\newcommand{\Xkk}{\mathcal{X}}
\newcommand{\xkk}{x}

\newcommand{\KL}[2]{D_{\mathrm{KL}} \left( {#1} \middle\| {#2} \right)}

\def\part#1#2{\frac{\partial #1}{\partial #2}}
\def\vec#1{\mbox{\boldmath $#1$}}

\begin{document}

\def\BibTeX{{\rm B\kern-.05em{\sc i\kern-.025em b}\kern-.08em
    T\kern-.1667em\lower.7ex\hbox{E}\kern-.125emX}}
\markboth{\journalname, VOL. XX, NO. XX, XXXX 2017}
{Author \MakeLowercase{\textit{et al.}}: Preparation of Papers for IEEE Control Systems Letters (August 2022)}

\title{Imitation-regularized Optimal Transport on Networks: Provable Robustness and Application to Logistics Planning}

\author{Koshi Oishi$^{a}$, Yota Hashizume$^{b}$, Tomohiko Jimbo$^{c,a}$, Hirotaka Kaji$^{c}$,~\IEEEmembership{Member,~IEEE}, and Kenji Kashima$^{b}$,~\IEEEmembership{Senior~Member,~IEEE,}
\thanks{$^{a}$ Toyota Central R\&D Labs., Inc., 41-1 Yokomichi, Nagakute, Aichi, Japan; \texttt{e1616@mosk.tytlabs.co.jp}}
\thanks{$^{b}$ Graduate School of Informatics, Kyoto University, Kyoto, Japan}%
\thanks{$^{c}$ Frontier Research Center, Toyota Motor Corporation, Aichi, Japan}%
}

\maketitle
\thispagestyle{empty}

\begingroup
  \renewcommand\thefootnote{}            % 脚注マークを出さない
  \footnotetext{\footnotesize
  © 2025 IEEE. Personal use of this material is permitted. 
  Permission from IEEE must be obtained for all other uses, in any current or future media, 
  including reprinting/republishing this material for advertising or promotional purposes, 
  creating new collective works, for resale or redistribution to servers or lists, 
  or reuse of any copyrighted component of this work in other works.}
  \addtocounter{footnote}{-1}            % カウンタを戻す
\endgroup

\begin{abstract}
Transport systems on networks are crucial in various applications, but face a significant risk of being adversely affected by unforeseen circumstances such as disasters.
The application of entropy-regularized optimal transport (OT) on graph structures has been investigated to enhance the robustness of transport on such networks.
In this study, we propose an imitation-regularized OT (I-OT) that mathematically incorporates prior knowledge into the robustness of OT.
This method is expected to enhance interpretability by integrating human insights into robustness and to accelerate practical applications. 
Furthermore, we mathematically verify the robustness of I-OT and discuss how these robustness properties relate to real-world applications.
The effectiveness of this method is validated through a logistics simulation using automotive parts data.
\end{abstract}


\section{Introduction}\label{sec:intro}
Transport systems on networks are crucial components that support modern society \cite{network}. 
However, these systems are at risk of being adversely affected by unforeseen circumstances \cite{CHOWDHURY2021102271, huang2022supply}.
Therefore, enhancing the robustness of these systems is crucial for society.
The optimal transport (OT) problem \cite{peyre2019computational, villani2021topics} is a type of optimization problem that considers mass transportation.
In addition, an efficient solution for OT with entropy regularization (also referred to as maximum entropy (MaxEnt) OT), proposed by Cuturi et al. \cite{cuturi2013sinkhorn}, has been widely applied to loss functions in deep learning.
In previous studies, it was demonstrated via the Schr\"{o}dinger bridge with a special stochastic process, that is, the Ruelle--Bowens (RB) random walk \cite{delvenne2011,parry1964intrinsic,ruelle2004thermodynamic}, that MaxEnt OT on graph structures can derive robust transport over networks \cite{chen2016robust, kk2022, oishi2023resilience}.
However, this robustness has not yet been mathematically discussed.
Furthermore, the robustness based solely on entropy is insufficient for applications in societies that are built upon diverse knowledge.

This study proposes imitation-regularized OT (I-OT), which mathematically incorporates prior knowledge such as human insights to achieve practical robustness.
In real-world operations, cost optimization is usually addressed first \cite{chopra2015supply}.
Therefore, handling transportation costs and prior knowledge as distinct parameters is practical because they are typically treated independently.
I-OT enhances interpretability in practical applications by preserving the independence of the cost optimization and imitation terms.
In addition, inspired by the robustness of entropy-regularized reinforcement learning \cite{eysenbach2021}, the robustness of I-OT is rigorously formulated and its relevance to real-world applications is discussed.
Finally, simulations based on real-world logistics data demonstrate that I-OT is suitable for practical applications.

The remainder of this paper is organized as follows: 
Section \ref{sec:OT} presents the OT problem and Schr\"{o}dinger bridge. 
In Sections \ref{sec:imi} and \ref{sec:robust}, we formulate I-OT and show its robustness. 
In Section \ref{sec:sim}, we examine the usefulness of the proposed method by applying it to logistics planning. 
Finally, concluding remarks are presented in Section \ref{sec:concl}.

%%%%%%%%%%%%%%%%%%%%%%%%%%%%%%%%%%%%%%%%%%%%%%%%%%%%%%%%%%%%%%%%%%%%%%%%%%%%%%
\section{Preliminary: optimal transport and Schr\"{o}dinger bridge over networks}
\label{sec:OT}
This section outlines OT and the Schr\"{o}dinger bridge. 

%%%%%%%%%%%%%%%%%%%%%%%%%%%%%%%%%%%%%%%%%%%%%%%%%%%%%%%%%%%%%%%%%%%%%%%%%%%%%%
\subsection{Optimal transport}
In this study, the time index is discrete and finite: $\mathcal{T} = \{0,\dots,T \}$. 
Let $\mathcal{G} (\mathcal{V},\mathcal{E})$ be a directed graph with a set of $n$ nodes $\mathcal{V} = \{1,\dots,n\}$ and edges $\mathcal{E} \subset \mathcal{V} \times \mathcal{V}$.
In addition, let $\eCkk : \mathcal{V} \times \mathcal{V} \rightarrow \mathbb{R}$ be a cost function, such that 
$\eCkk(i,j) =\infty$ for $(i,j)\notin \mathcal{E}$.
% This corresponds to the adjacency matrix of the graph $\mathcal{G}$. The network consisting of the cost function $C$ and directed graph $\mathcal{G}$ is denoted by $\mathcal{G}(C)$.
Subsequently, let $\nu_0$ and $\nu_T$ be the probability distributions of $\mathcal{V}$. Then, we consider the transition from the initial distribution $\nu_0$ to the terminal distribution $\nu_T$ on $\mathcal{G}$ in $T$ steps.
For instance, $\nu_0(1) = 1$ indicates that the initial distribution of the transport targets is concentrated at node 1. 
By denoting the transitions in the network as $x := (x_0,\dots,x_T)\in \Xkk := \mathcal{V}^{T+1}$, we define its cost as follows:
\begin{equation}\label{eq:decompC}
    \Ckk :=\sum_{t=0}^{T-1} \eCkk(x_t, x_{t+1}).
\end{equation}
We can formulate the OT problem in the network as follows:
\begin{align}
    \begin{split}
        & \min_{P} \sum_{\xkk \in \Xkk} \Ckk P(\xkk), \\[4pt]
        & {\rm subject \:\: to} \:\: P(x_0) = \nu_0 (x_0) ,\\
        & \hspace{16.8mm} P(x_T) = \nu_T (x_T),
    \end{split}
    \label{eq:ot}
\end{align}
where decision variable $P(\cdot)$ is the probability distribution of $\Xkk$. 
Specifically, $P(x)$ assigns a probability to each sequence $x \in \Xkk$.
We define the marginal distribution of each $x_0 \in \mathcal{V}$ as follows:
\begin{align*}
 P(x_0) := \sum_{(x_1, \dots, x_T) \in \mathcal{V}^T} P\left(x = \left(x_0, \dots, x_T \right) \right).
\end{align*}
The terminal marginal distribution $P(x_T)$ is defined in a similar manner.
This problem involves linear programming with a maximum of $n^{T+1}$-dimensional decision variables. 
In general, solving this optimization problem directly is challenging because we are often interested in complex and large networks.

%%%%%%%%%%%%%%%%%%%%%%%%%%%%%%%%%%%%%%%%%%%%%%%%%%%%%%%%%%%%%%%%%%%%%%%%%%%%%%%
\subsection{Schr\"{o}dinger bridge}\label{sec:sb}
% Cuturi  et al. used entropy regularization as a fast method for solving optimal transport problems

Subsequently, we introduce the Schr\"{o}dinger bridge \cite{sch, chen2016robust,chen2017efficient}. 
Let $\mathfrak{M}$ be a Markovian probability distribution on $\Xkk$ that is expressed as follows: 
\begin{align}
    \mathfrak{M}(\xkk) = \mu_0(x_0) \bm{M}_{x_0,x_1} \: \dots \: \bm{M}_{x_{T-1},x_T},  \label{eq:prior_distribution}
\end{align}
where $\mu_0$ is a probability distribution on $\mathcal{V}$ that serves as the initial distribution for $\mathfrak{M}$ and $\bm{M}\in\mathbb{R}^{n\times n}$ is an irreducible time-invariant transition matrix.
The Schr\"{o}dinger bridge can be formulated as follows:
\begin{align}
    \begin{split}
        & \min_{P} \KL{P}{\mathfrak{M}} ,\\
        & {\rm subject \:\: to} \:\: P(x_0) = \nu_0 (x_0) ,\\
        & \hspace{16.8mm} P(x_T) = \nu_T (x_T). 
    \end{split}
    \label{eq:sch}
\end{align}
The Schr\"{o}dinger bridge aims to determine the time-evolving probability distribution $P$ that satisfies a fixed initial distribution $\nu_0$ and the terminal distribution $\nu_T$ that is closest to the prior distribution $\mathfrak{M}$. 
This problem is known to have a unique solution \cite[Theorem 3]{georgiou2015schrodinger}.
We consider two non-negative functions $\varphi$ and $\hat{\varphi}$ on $\mathcal{T}\times\mathcal{V}$ with the boundary constraints $\nu_0$ and $\nu_T$. 
These are required to satisfy
\begin{align}
    \begin{split}
        \varphi(t,i) &= \sum_{j \in \mathcal{V}} \bm{M}_{i,j} \varphi(t+1,j) ,\\
        \hat{\varphi}(t+1,j) &= \sum_{i \in \mathcal{V}} \bm{M}_{i,j} \hat{\varphi}(t,i)
    \end{split}
    \label{eq:phi_t}
\end{align}
and the boundary conditions
\begin{align}
    \begin{split}
        \varphi(0,x_0) \cdot \hat{\varphi}(0,x_0) &= \nu_0(x_0) \:\:\:\: \forall{x_0} \in \mathcal{V} ,\\[4pt]
        \varphi(T,x_T) \cdot \hat{\varphi}(T,x_T) &= \nu_T(x_T) \:\:\:\: \forall{x_T} \in \mathcal{V}
    \end{split}
    \label{eq:phi_b}
\end{align}
for all $t \in \mathcal{T}$.
Using $\varphi$, we define the transition matrix $\Pi(t)$ for each $t\in \{0,\dots,T-1\}$ as follows:
\begin{align}
    \Pi (t) := {\rm diag}(\varphi(t))^{-1} \bm{M} {\rm diag}(\varphi (t+1)).
    \label{eq:pi}
\end{align}
Then, the probability distribution on $\Xkk$ is obtained as
\begin{align}
\label{eq:Pstar}
    \begin{split}
        & P^*[\nu_0, \nu_T](x_0, \dots, x_T) = \\[4pt]
        & \hspace{10mm} \nu_0(x_0) \Pi (0)_{x_0,x_1} \:\: \dots \:\: \Pi(T-1)_{x_{T-1},x_T}.
    \end{split}
\end{align}
\begin{proposition}
\label{prop:schr}
    If all components of $\bm{M}^{T}$ are positive, there exists a unique pair of non-negative functions $\varphi, \hat{\varphi}$ that satisfies \eqref{eq:phi_t} and \eqref{eq:phi_b}.
    Moreover, the probability distribution $P^*[\nu_0,\nu_T]$ defined by \eqref{eq:Pstar} is the unique solution to \eqref{eq:sch}.
\end{proposition}
This result yields the Sinkhorn iteration, which is an effective algorithm for solving the Schr\"{o}dinger bridge \cite{chen2016robust}. 
We denote $\varphi_0$ as the vector defined as $\varphi_0: =[ \varphi(0,1),\dots,\varphi(0,n)]^\top \in \mathbb{R}^n$.
Vectors $\varphi_T,\hat{\varphi}_0,$ and $\hat{\varphi}_T$ are defined in a similar manner.
As the recursive relations in \eqref{eq:phi_t} can be consolidated using $\bm{M}^T$, \eqref{eq:phi_t} and \eqref{eq:phi_b} are expressed as follows:
\begin{equation}\label{eq:sch-system}
    \begin{split}
    &\varphi_0 = \bm{M}^T\varphi_T, \\
    &\hat{\varphi}_T = (\bm{M}^T)^\top \hat{\varphi}_0, \\
    &\varphi_0 \odot \hat{\varphi}_0 = \nu_0, \\
    &\varphi_T \odot \hat{\varphi}_T = \nu_T, 
    \end{split}
\end{equation}
where $\odot$ is the Hadamard product.
The solution to $\varphi_0,\varphi_T,\hat{\varphi}_0,$ and $\hat{\varphi}_T$ can be rapidly obtained through the iteration of \eqref{eq:sch-system} \cite{chen2016robust}.
Based on the obtained solution and \eqref{eq:phi_t}, we compute $\varphi(t),\hat{\varphi}(t)$ for $t=1,...,T-1$. Consequently, the transition matrix $\Pi(t)$ is calculated using \eqref{eq:pi}. 

%%%%%%%%%%%%%%%%%%%%%%%%%%%%%%%%%%%%%%%%%%%%%%%%%%%%%%%%%%%%%%%
\section{Imitation-regularized optimal transport}
\label{sec:imi}
In this section, we present the proposed I-OT and clarify its relationship with MaxEnt OT and the Schr\"{o}dinger bridge.

%%%%%%%%%%%%%%%%%%%%%%%%%%%%%%%%%%%%%%%%%%%%%%%%%%%%%%%%%%%%%%
\subsection{Formulation of I-OT}
We introduce the probability distribution $Q$ on $\Xkk$ as the target of imitation, as follows:
\begin{align}
    Q(\xkk) = \mu_{Q_0}(x_0)(\bm{R}_Q)_{x_0,x_1} \: \dots \: (\bm{R}_Q)_{x_{T-1}, x_T},\label{eq:imit_distribution}
\end{align}
where $\bm{R}_Q$ is a time-invariant stochastic matrix and $\mu_{Q_{0}}$ denotes the initial distribution for $Q$.
I-OT, which imitates $Q$, is expressed as follows:
\begin{prob}[I-OT]
    \label{prob:imi}
    Given the cost function $C$ and probability distribution $Q$ on $\Xkk$ and $\alpha>0$, find
    \begin{align}
        \begin{split}
            & \min_{P} \sum_{\xkk \in \mathcal{X}} \Ckk P(\xkk) + \alpha \KL{P}{Q} ,\\[4pt]
            & {\rm subject \:\: to} \:\: P(x_0) = \nu_0 (x_0) ,\\
            & \hspace{16.8mm} P(x_T) = \nu_T (x_T).
        \end{split}
        \label{eq:imi}
    \end{align}
\end{prob}
The parameter $\alpha$ represents the weight of imitation.
The obtained solution is close to $Q$ if $\alpha$ is large, and approaches the solution for OT in \eqref{eq:ot} if $\alpha$ is small.
In practical logistics operations, transportation plans are often determined based on prior transportation experience.
I-OT provides a framework for intuitively incorporating this experience as $Q$.
\begin{remark}
\label{re:IOTEOT}
    MaxEnt OT is formulated as
    \begin{align}
        \begin{split}
            & \min_{P} \sum_{\xkk \in \mathcal{X}} C(x) P(\xkk) - \alpha \mathcal{H}(P),\\[4pt]
            & {\rm subject \:\: to} \:\: P(x_0) = \nu_0 (x_0) ,\\
            & \hspace{16.8mm} P(x_T) = \nu_T (x_T),
        \end{split}
        \label{eq:MaxEntOT}
    \end{align}
    where the Shannon entropy of $P$ is defined as
    \begin{align}
        \mathcal{H}(P) := -\sum_{x \in \mathcal{X}} P(\xkk) \log{P(\xkk)}.
        \label{eq:entropy}
    \end{align}
    We can show that Problem~\ref{prob:imi} is equivalent to MaxEnt OT whose cost is modified as
     \begin{align}
        C_Q(x) := C(x) -\alpha \log{Q(x)}.
        \label{eq:Cq}
    \end{align}
    When $Q$ is a uniform distribution, the solution of Problem~\ref{prob:imi} is equivalent to that of \eqref{eq:MaxEntOT}.
\end{remark}
Therefore, I-OT is expected to exhibit not only the cost optimality and imitation properties but also the robustness characteristic of MaxEnt OT.

%%%%%%%%%%%%%%%%%%%%%
\subsection{Relation to Schr\"{o}dinger bridge}
MaxEnt OT is known to be equivalent to the Schr\"{o}dinger bridge \cite{Estimating_sch, leonard2013survey, mutli_sch}.
When dealing with transportation on network structures, the RB random walk \cite{delvenne2011} is applied.
The prior distribution $\mathfrak{M}_{\rm RB}$ based on the RB random walk is derived using the following matrix, which is defined by the cost functions $c$ and $\alpha > 0$:
\begin{align}
        \bm{B}_{i,j} := \exp(- \frac{\eCkk(i,j)}{\alpha} ).
\end{align}
Note that the graph $\mathcal{G}$ is assumed to be strongly connected and aperiodic.
In this case, all components of $\bm{B}$ are non-negative and a natural number $k$ exists such that all components of $\bm{B}^k$ are positive.
Therefore, from the Perron--Frobenius theorem, the spectral radius $\lambda_B$ of $\bm{B}$ is a simple eigenvalue of $\bm{B}$, and the components of the left eigenvector $\bm{u} \in \mathbb{R}^n$ and right eigenvector $\bm{v} \in \mathbb{R}^n$ are all positive.
\begin{align*}
    \begin{split}
        \bm{B}^{\top} \bm{u} &= \lambda_B \bm{u} ,\\
        \bm{B}\bm{v} &= \lambda_B \bm{v} .
    \end{split}
\end{align*}
In this case, $\bm{u}$ and $\bm{v}$ are normalized to satisfy 
\begin{align*}
    \sum_{i=1}^{n} \bm{u}_i \bm{v}_i = 1.
\end{align*}
Thus, the function $\bm{v}_{\rm RB}(i) = \bm{u}_i \bm{v}_i$ on $\mathcal{V}$ is a probability mass function.
The transition matrix is defined as follows:
\begin{align*}
    \bm{R} := \lambda_B^{-1} {\rm diag} (\bm{v})^{-1} \bm{B} {\rm diag} (\bm{v}).
\end{align*}
In this case, $\mathfrak{M}_{\rm RB}$, which is defined below, is a probability distribution on $\Xkk$ and is referred to as an RB random walk.
\begin{align}
    \mathfrak{M}_{\rm RB} (\xkk) := \bm{v}_{\rm RB} (x_0) \bm{R}_{x_0,x_1} \:\: \dots \:\: \bm{R}_{x_{T-1}, x_T}.
    \label{eq:RB}
\end{align}
The RB random walk is a probability distribution (on paths) with the highest entropy rate under a given cost.
Thus, within this cost setting, it can be regarded as the most uniform distribution.
The relationship between I-OT and the Schr\"{o}dinger bridge can be derived from a prior distribution that is defined as follows:
\begin{align}
    \mathfrak{M}_Q(\xkk) := \mathfrak{M}_{\rm RB}(\xkk) Q(\xkk).
\label{eq:MQ}
\end{align}
\begin{thm}
\label{prop:imi}
    The optimal solution of \eqref{eq:sch} with $\mathfrak{M}_{Q}$ as its prior distribution is equivalent to that of Problem \ref{prob:imi}.
\end{thm}
\begin{pf}
    By omitting constant terms, \eqref{eq:sch} and \eqref{eq:MaxEntOT} hold the following relationship \cite[Remark 1]{Estimating_sch}:
    \begin{align}
        \KL{P}{\mathfrak{M}_{\rm RB}} = \frac{1}{\alpha} \left(\sum_{x \in \mathcal{X}} C(x)P(x) - \alpha \mathcal{H}(P) \right).
    \end{align}
    Consequently, we obtain 
    \begin{align}
         \begin{split}
            \KL{P}{\mathfrak{M}_{Q}} &= \KL{P}{\mathfrak{M}_{\rm RB}} - \sum_{x \in \mathcal{X}} P(x) \log{Q(x)}\\
            &= \frac{1}{\alpha} \left(\sum_{x \in \mathcal{X}} P(x) C(x) + \alpha \KL{P}{Q}\right) .
        \end{split}
        \label{eqa:sch_p}
    \end{align}
\end{pf}
\begin{remark}
    In contrast to $\mathfrak{M}_{\rm RB}$, $\mathfrak{M}_{Q}$ is not a probability density function for paths in general. The corresponding (normalized) probability density function is expressed as follows:
    \begin{equation}
        \overline{\mathfrak{M}}_Q( \xkk) 
        := \frac{\mathfrak{M}_Q( \xkk )}{\sum_{\xkk \in \mathcal{X}}\mathfrak{M}_Q(\xkk)} ,
    \end{equation}
    which satisfies 
    \begin{equation}
        \KL{P}{\mathfrak{M}_Q} = \KL{P}{\overline{\mathfrak{M}}_Q} + \sum_{\xkk \in \mathcal{X}}\mathfrak{M}_Q(\xkk). 
    \end{equation}
    The normalization of $\mathfrak{M}_{Q}$ is not required to determine the optimal $P$ because the second term on the right-hand side is independent of $P$. 
\end{remark}
Theorem~\ref{prop:imi} indicates that Problem \ref{prob:imi} can leverage the Schr\"{o}dinger bridge framework in \eqref{eq:sch-system}.
This framework can be solved by iterating the matrix operation in $n \times n$ dimensions, which is considerably faster than solving a linear programming problem with $n^{(T+1)}$ decision variables.
In practice, the derivation of $\bm{M}$ in \eqref{eq:sch-system} is crucial for computing I-OT within the Schr\"{o}dinger bridge framework.
$\mathfrak{M}_{Q}$ can be represented as follows:
\begin{align*}
    & \mathfrak{M}_{Q} = \mathfrak{M}_{\rm RB}(\xkk) Q(\xkk) = \\
        & \:\:\: v_{\rm RB}(x_0) \mu_{Q_0}(x_0)(\bm{M}_Q)_{x_0,x_1} \:\: \dots \:\: (\bm{M}_Q)_{x_{T-1}, x_T} ,\\[4pt]
        &\bm{M}_Q := \bm{R} \odot \bm{R}_Q.
\end{align*}
Therefore, computation is possible by simply setting $\bm{M}$ to $\bm{M}_Q$.
Note that the initial distribution of the prior distribution $\mu_{Q_0}$ is not required to derive the solution.

%%%%%%%%%%%%%%%%%%%%%%%%%%%%%%%%%%%%%%%%%%%%%%%%%%%%%%%%%%%%%%%%%%%%%%%%%%%
\section{Provable robustness}
\label{sec:robust}
In this section, the robustness of Problem~\ref{prob:imi} is clarified.
I-OT is expected to possess certain robustness owing to the equivalence with the Schr\"{o}dinger bridge involving the random walk, as demonstrated in Theorem~\ref{prop:imi}.
Therefore, we consider robust OT using a set of cost variations \cite{eysenbach2021}.
Defining the set of cost variations enables the acquisition of a robust solution by minimizing the cost in the worst-case scenario. 
The robust optimization problem for the set of cost functions $\mathcal{C}$ is as follows:
\begin{prob}[Robust OT]
\label{prob:robust}
    Given a robust cost function set $\mathcal{C}$, find 
    \begin{align}
        \begin{split}
            & \min_P \max_{\Tilde{C} \in \mathcal{C}} \sum_{\xkk \in \mathcal{X}} \tCkk P(\xkk) ,\\[4pt]
            & {\rm subject \:\: to} \:\: P(x_0) = \nu_0 (x_0) ,\\
            & \hspace{16.8mm} P(x_T) = \nu_T (x_T).
        \end{split}
        \label{eq:robust}
    \end{align}
\end{prob}
Problem \ref{prob:robust} involves deriving the optimal transportation plan for the worst cost from the set of cost functions $\mathcal{C}$.
As opposed to Problem~\ref{prob:imi} that includes regularization, the objective function of Problem~\ref{prob:robust} reflects the total cost under the worst-case scenario, and thus, has practical economic utility.
\begin{thm}
    \label{thm:robust}
    Supposing that the robust cost function set is expressed as follows:
    \begin{align}
        \begin{split}
            \mathcal{C} := \left\{\Tilde{C}  : \alpha \log \left( \sum_{ \xkk \in \mathcal{X}} Q(\xkk)  \right. \right. \hspace{-1mm} \exp{\frac{\tCkk-\Ckk}{\alpha}}\Biggr)  \leq \epsilon \Biggr\},
        \end{split}
        \label{eq:robust_set}
    \end{align}
    Problems~\ref{prob:imi} and~\ref{prob:robust} have the same objective function up to an additive constant, which is expressed as follows:
    \begin{align}
        \begin{split}
            \max_{\Tilde{C} \in \mathcal{C}} \sum_{x \in \calX}  \Tilde{C}(x) P(x) =  \sum_{x \in \calX} P(x) C(x) + \alpha \KL{P}{Q} + \epsilon.
        \end{split}
    \label{eq:appB}
    \end{align}
\end{thm}
\begin{pf}
Similar to the approach in \cite{eysenbach2021}, applying the Karush--Kuhn--Tucker conditions to Problem~\ref{prob:robust} with \eqref{eq:robust_set} yields the following:
\begin{align}
  \Tilde{C}(x) = C(x) - \alpha \log{\left(\frac{Q(x)}{P(x)}\right)}  +\epsilon .
  \label{eq:TCeps}
\end{align}
Thus, \eqref{eq:appB} is obtained by substituting \eqref{eq:TCeps} into \eqref{eq:robust}.
\end{pf}
\begin{remark}\label{rm:eps}
    Although the set in \eqref{eq:robust_set} is dependent on $\epsilon$, \eqref{eq:appB} shows that Problem \ref{prob:imi} is independent of $\epsilon$. 
    Thus, when the robust set is \eqref{eq:robust_set}, Problem \ref{prob:robust} has the same solution regardless of $\epsilon$. 
    However, the optimal value of Problem \ref{prob:robust} is greater than that of Problem \ref{prob:imi} by $\epsilon$.
    Replacing the Shannon entropy in \eqref{eq:entropy} with the Tsallis entropy leads to the optimal solution depending on $\epsilon$ \cite{hashizume2024tsallis}.
\end{remark}
Theorem~\ref{thm:robust} extends Theorem~4.1 in~\cite{eysenbach2021} by introducing a discrete setting, incorporating the cost in~\eqref{eq:Cq}, and clarifying the objective function.
The robust set in~\eqref{eq:robust_set} involves a cumulant generating function that captures the mean, variance, and skewness and is employed in entropy-based risk robust optimization methods~\cite{glasserman2014robust}. 
Furthermore, $\mathcal{C}$ is convex.

From a practical perspective, $Q(x)$ can be interpreted as a measure of desirability for a path $x$, which is implicitly constructed based on the past transportation experience. 
In particular, paths $x$ where $Q(x)$ is zero are not selected in I-OT due to the definition of the KL divergence. 
Note that, using the log-sum-exp approximation, the set in \eqref{eq:robust_set} can be approximated as 
\begin{align}
        \begin{split}
            \mathcal{C} \approx  \left\{\tilde{C}  : \max_x \left[\Delta C(x) + \alpha \log \left( Q(x) \right) \right] \le \epsilon \right\}
        \end{split}
        \label{eq:robust_set2}
\end{align}
for large $\alpha$, where $\Delta C(x) := \tilde{C} (x) - C(x)$. 
Therefore, $\epsilon$ is related to the maximum allowable value for the sum of cost fluctuation and desirability. 
This indicates for a fixed $\epsilon$ that \emph{small} fluctuations are only allowed for desirable paths, and that paths with large expected fluctuations must have a \emph{low} level of desirability. 

Theorem \ref{thm:robust} can also be used to estimate the worst-case performance for a fixed transport plan $P$. Suppose that a real cost set $\mathcal{C}_{\rm real}$ is available. 
Then, it seems reasonable to choose
\begin{align}\label{eq:eps_real}
    \epsilon_{\rm real} := \max_{x\in \mathcal{X},\tilde C\in \mathcal{C}_{\rm real}} \left[\Delta C(x) + \alpha \log \left( Q(x) \right) \right]
\end{align}
and estimate the worst-case performance by
\begin{align}
    \begin{split}
        \max_{\tilde{C} \in \mathcal{C}_{\rm real}} &\sum_{x \in \calX}  \Tilde{C}(x) P(x) \approx   \\
        &\sum_{x \in \calX} P(x) C(x) + \alpha \KL{P}{Q} + \epsilon_{\rm real}.
    \end{split}
\end{align}
Although calculating $\epsilon_{\rm real}$ in \eqref{eq:eps_real} accurately is challenging, a practical strategy is to extract several candidate paths $x$ that are expected to have large $Q(x)$ or large maximum fluctuation $\max_{\tilde{C} \in \mathcal{C}_{\rm real}}\Delta C(x)$, and then compute the maximum of $\Delta C(x) + \alpha \log \left( Q(x) \right)$ over only these paths. 

%バグ対策
\textcolor{black}{}
%%%%%%%%%%%%%%%%%%%%%%%%%%%%%%%%%%%%%%%%%%%%%%%%%%%%%%%%%%%%%%%%%%%%%%%%%%%%
\section{Application to logistics}
\label{sec:sim}

We performed numerical calculations to demonstrate applications in logistics using automotive parts delivery data from Japan. 
In this demonstration, we verified the I-OT robustness of \eqref{eq:robust_set}, assuming a disaster scenario with a Mt. Fuji eruption.

\begin{figure}[t]
        \vspace{2mm}
	\centering
	\subfigure [Location of each base in Japan]{
            \includegraphics[width=4.1cm, trim= 0.1cm 0 1.3cm 0.2cm,clip]{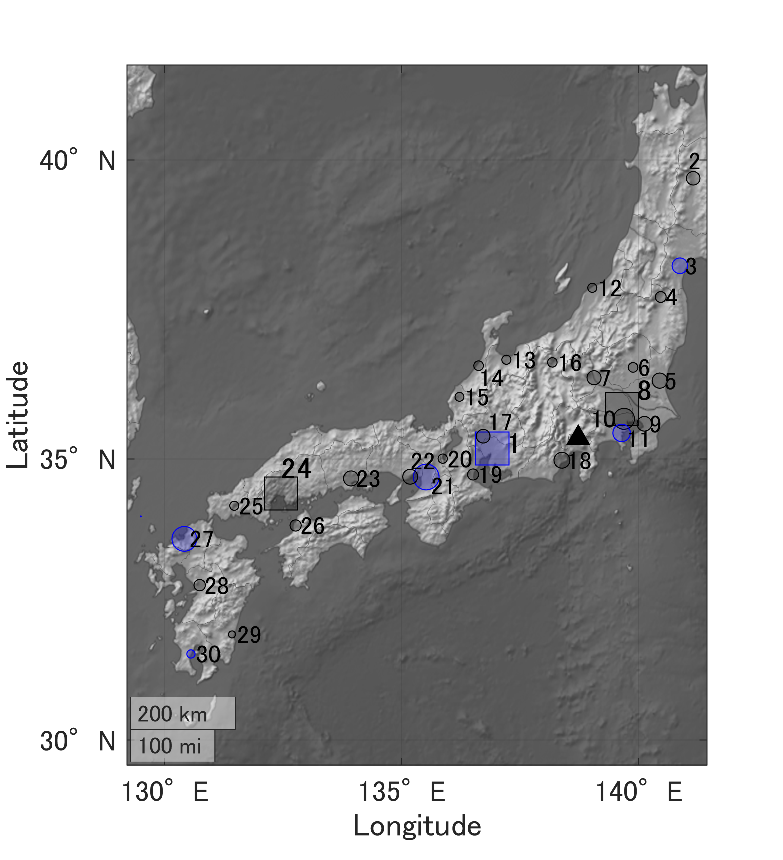}
            \label{fig:map}}
	\subfigure [Network]{
        \includegraphics[width=4.1cm, trim= 0.1cm 0 1.2cm 0.2cm,clip]{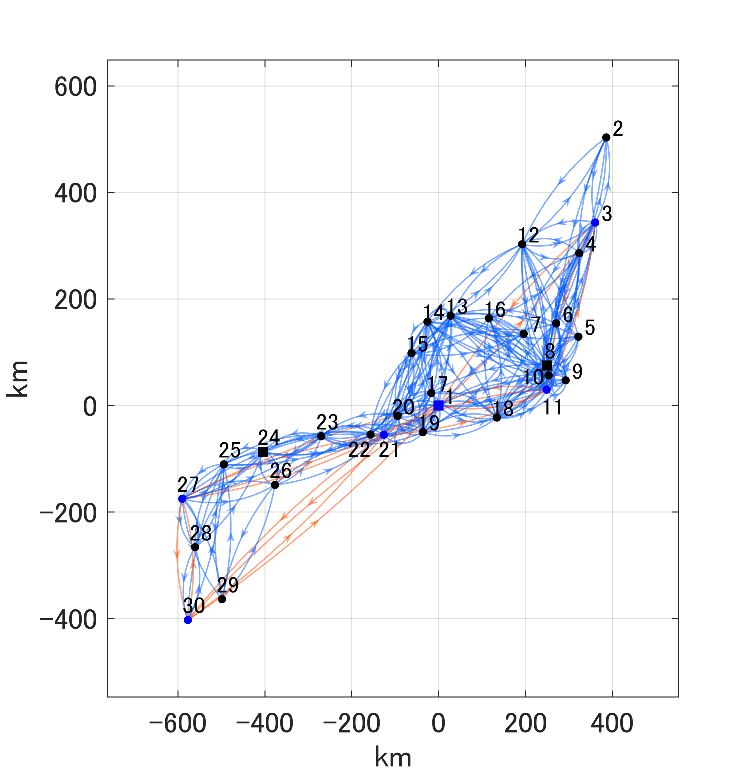}
		\label{fig:net}}
	\caption{Automotive parts transportation. (a) Node locations and magnitudes of supply and demand. The circular and square nodes denote the delivery destinations and supply sources, respectively. The blue nodes denote ports and the triangular node denotes Mt. Fuji. The size of each node represents the supply and demand quantities. (b) Digraph representation. The axes denote the position (units: km). The blue and red edges denote roads and maritime paths, respectively. The self-loop representing storage is omitted.}
	\label{fig:tnet}
\end{figure}

%%%%%%%%%%%%%%%%%%%%%%%%%%%%%%%%%%%%%%%%%%%%%%%%%%%%%%%%%%%%%%%%%%%%%%%%
\subsection{Formulation}
The total number of nodes was $n=30$ (Hokkaido and Okinawa were excluded).
We assumed that the supply sites for the parts were located in Aichi ($1$), Saitama ($8$), and Hiroshima ($24$), Japan, which are known for their high automotive parts production. 
The node locations are shown in Fig.~\ref{fig:map}.
The supply quantity for each supply node was allocated by dividing the demand into three equal parts, ensuring that no fractions were generated:
\begin{align*}
    \nu_0(x_0) = \left \{
    \begin{array}{ll}
          \frac{490}{q_{\rm total}}  & x_0 \in \{1,8\}  ,\\[3pt]
          \frac{489}{q_{\rm total}}  & x_0 = 24,  \\
           0    & \mbox{\rm others} ,\\
    \end{array}
    \right.
\end{align*}
where $q_{\rm total} = 1469$ is the total demand.
The demand was characterized by $\nu_T$, where all nodes except for the supply node had a positive demand (Fig.~\ref{fig:map}). 

To leverage prior knowledge for robustness, we included low-cost edges that are vulnerable to disasters and high-cost edges that are resistant to them.
Figure~\ref{fig:net} depicts the overall network.
This network comprised three types of edges:
\begin{itemize}
    \item Road: These were the roads connecting major cities.
    \item Maritime: These edges represented routes connecting major ports ($1$, $3$, $11$, $21$, $27$, and $30$) and incurred higher costs than land transportation.
    \item Storage: These were the self-loop edges, indicating the storage of parts. These self-loops were charged storage costs. 
\end{itemize}
The cost for the road edges was based on the Euclidean distance, whereas the quadruple of the Euclidean distance was considered for the maritime edges. 
Storage, which was represented by self-loop edges, was allocated a cost corresponding to a Euclidean distance of 10 km.

\begin{figure}[t]
        \vspace{2mm}
	\centering
	\subfigure [Expected damaged edges under disaster of Mt.~Fuji.]{
            \includegraphics[width=4.1cm, trim= 0.1cm 0 1.2cm 0.2cm,clip]{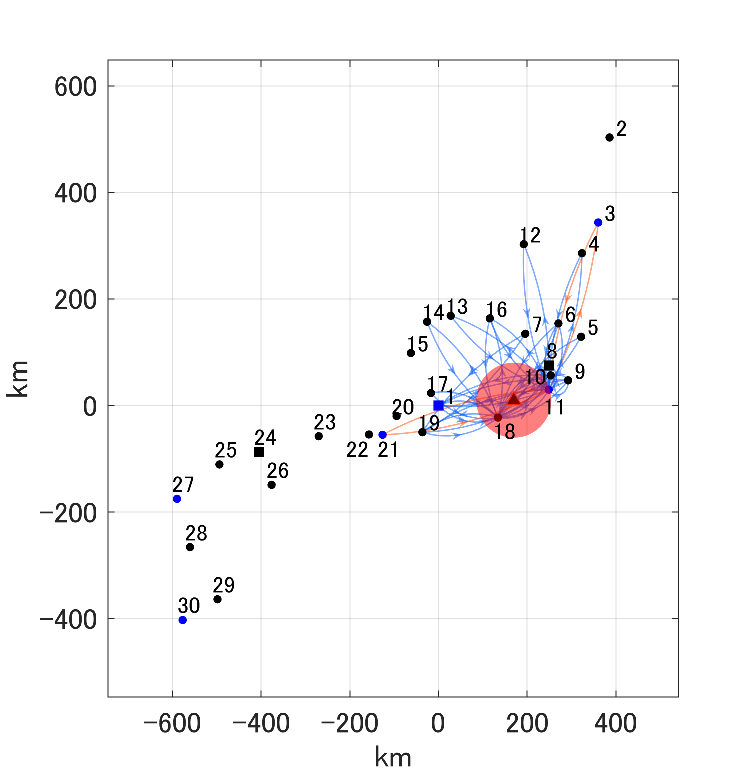}
            \label{fig:map_disaster}}
	\subfigure [$\bm{R}_{Q}$]{
        \includegraphics[width=4.1cm, trim= 0.1cm 0 1.2cm 0.2cm,clip]{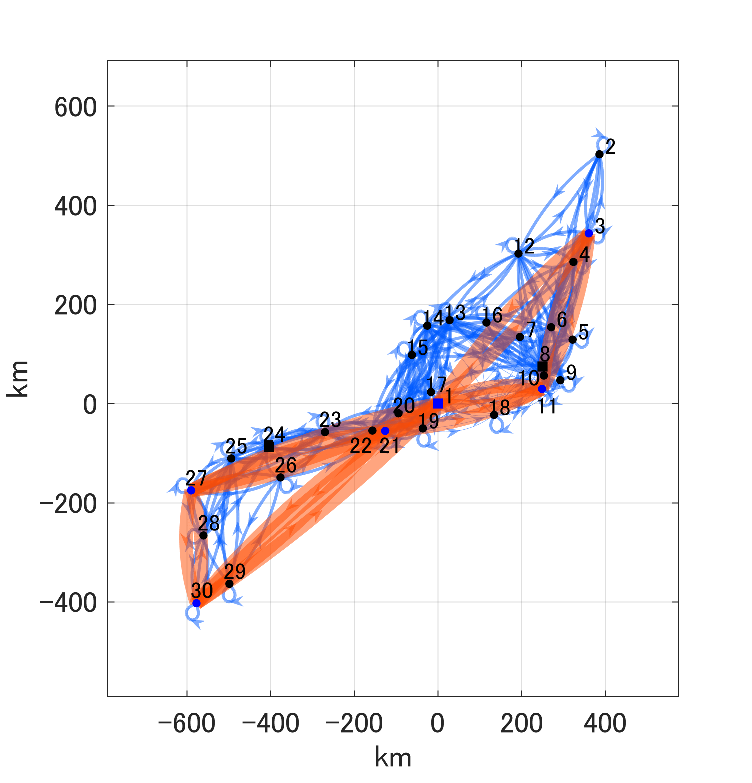}
		\label{fig:s2_q}}
	\caption{A priori risk information. The edge width is not an exact value as visibility is prioritized.}
	\label{fig:disaster}
\end{figure}

Figure~\ref{fig:map_disaster} depicts the expected affected edges for a Mt. Fuji eruption.
We incorporated this information into $Q$ in \eqref{eq:robust_set}, treating it as prior knowledge. 
Transition matrix $\bm{R}_{Q}$ from \eqref{eq:imit_distribution} was used to realize $Q$ because the prior knowledge was provided in the form of edges.
Based on this prior knowledge, $\bm{R}_{Q}$ was expressed as follows: 
\begin{align}
    [\bm{R}_{Q}]_{i,j} = \left \{
    \begin{array}{ll}
         10^{-5} &  \mbox{\rm for expected affected edges}, \\
         10^{5}  & \mbox{\rm for marine  transport},  \\
         0    & \mbox{for }(i,j) \notin \mathcal{E}, \\
         1    & \mbox{others }.
    \end{array} \right. 
    \label{eq:s1_q}
\end{align}
Section~\ref{sec:robust} with \eqref{eq:imit_distribution} indicates that larger values of $\bm{R}_{Q}$ correspond to less fluctuation.
Thus, prior knowledge was incorporated into the OT to reflect significant cost fluctuations in roads near Mt. Fuji and minimal fluctuations in maritime routes.
Figure~\ref{fig:s2_q} shows $\bm{R}_Q$, where the thicker edges indicate less impact from the disaster.
The red edges representing maritime routes are thicker, whereas the blue edges representing roads near Mt. Fuji are thinner.

\begin{comment}
\begin{figure}[t]
	\centering
		\includegraphics[keepaspectratio, width=6cm]{pic/Fuji.eps}
	  \caption{Damaged edges under disaster of Mt.~Fuji.}\label{fig:map_disaster}
\end{figure}

\begin{figure}[t]
	\centering
		\includegraphics[keepaspectratio, width=6cm]{pic/s2_Q2.eps}
	  \caption{A priori risk information $\bm{R}_{Q}$. The edge width is not an exact value as the visibility is prioritized.}\label{fig:s2_q}
\end{figure}
\end{comment}

\begin{comment}
\begin{figure*}[t]
	\centering
		\includegraphics[keepaspectratio, width=15cm, trim= 3.2cm 0 3.6cm 0,clip]{pic/s2_P2.eps}
	  \caption{Robust optimal logistics plan $P_{\rm I}$. The maritime edges are emphasized.} \label{fig:s2_time}
\end{figure*}
\end{comment}
\begin{figure*}[t]
        \vspace{2mm}
	\centering
	\subfigure [$P_{\rm I}$]{
            \includegraphics[width=15cm, trim= 3.2cm 0 3.6cm 0.2cm,clip]{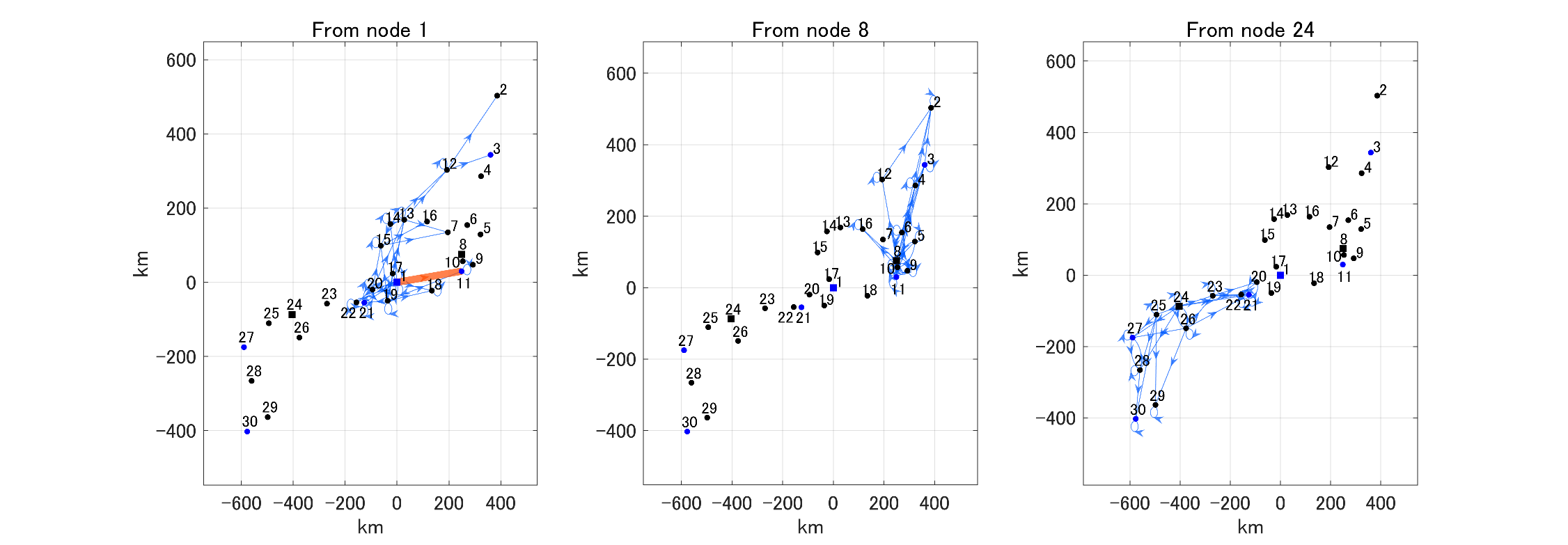}
            \label{fig:s2_IOT}} \\
	\subfigure [$P_{\rm MAX}$]{
        \includegraphics[width=15cm, trim= 3.2cm 0 3.6cm 0,clip]{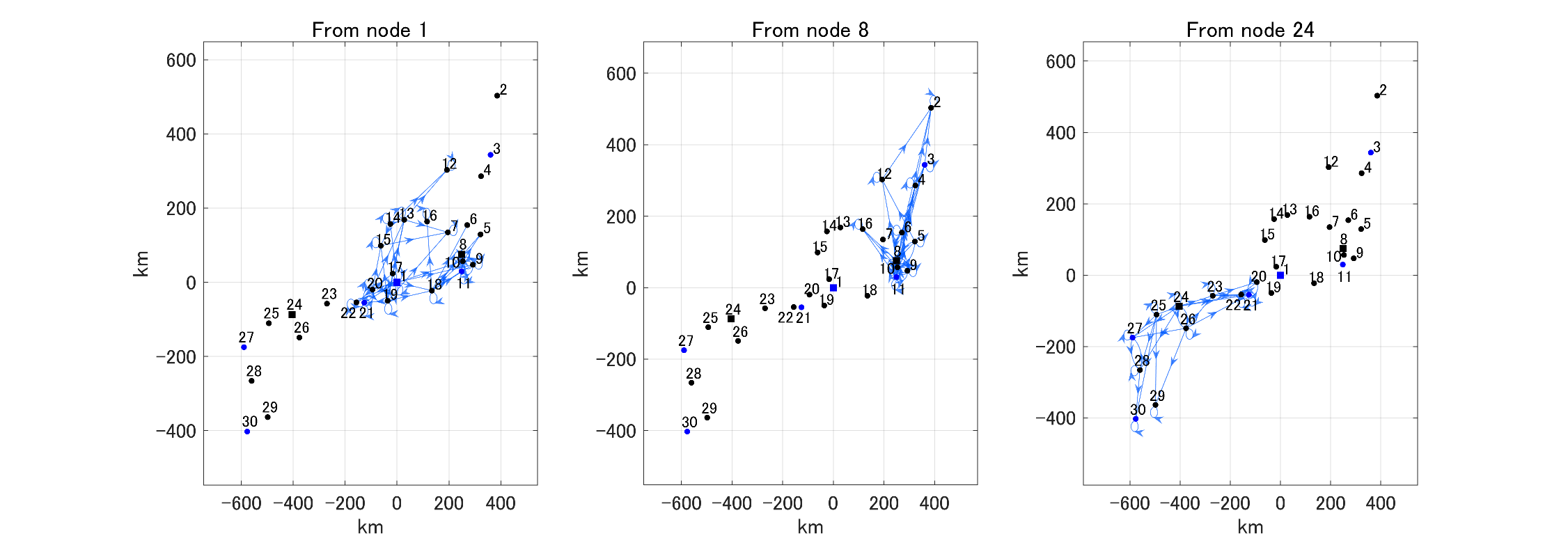}
		\label{fig:s2_EOT}}
	\caption{Logistics plan. The maritime edges are emphasized with thicker lines.}
	\label{fig:s2_time}
\end{figure*}

%バグ対策
%\textcolor{black}{}
%%%%%%%%%%%%%%%%%%%%%%%%%%%%%%%%%%%%%%%%%%%%%%%%%%%%%%%%%%%%%%%%%%%%%%
\subsection{Results}\label{sec:Fuji}
We present the numerical results for the proposed I-OT method and compare them with those of existing approaches (MaxEnt OT and the classical OT).
Figure~\ref{fig:s2_time} presents the I-OT solution $P_{\rm I}$ and standard MaxEnt OT solution $P_{\rm MAX}$ under $\alpha=30$ and $T=3$.
Compared with $P_{\rm MAX}$, the strategy of $P_{\rm I}$ employed marine routes that were expensive but robust.
Furthermore, $P_{\rm I}$ used fewer road routes near Mt. Fuji than $P_{\rm MAX}$.
This indicates that $P_{\rm I}$ is robust based on prior knowledge, in contrast to $P_{\rm MAX}$, which provides uniform robustness.

Assuming a 10-fold increase in the cost of half of the predicted edges damaged by the disaster in Fig.~\ref{fig:map_disaster}, we compared the cost of I-OT, MaxEnt OT, and OT.
To solve OT, we used SeDuMi in CVX, which is a package for specifying and solving convex programs \cite{sturm1999,cvx,gb08}.
The total cost is summarized in Table~\ref{tb:s2_tloss}. 
$P_{\rm OT}$ are the solutions to OT.
I-OT significantly reduced the costs after the disaster, although it incurred slightly higher costs beforehand.
This demonstrates that the transport plan derived from I-OT is less vulnerable to disaster impacts.
Although MaxEnt OT, which introduces robustness only through entropy, performed better than OT in disaster scenarios, I-OT, with its ability to incorporate prior knowledge, performed significantly better than MaxEnt OT.
These results demonstrate that I-OT effectively utilizes prior knowledge.

\begin{table}[t]
	\caption{Total transportation costs}
	\label{tb:s2_tloss}
	\centering
	\begin{tabular}{ccc|ccc}
			\hline
			    \multicolumn{3}{c|}{Before disaster} & \multicolumn{3}{c}{After disaster} \\
               $\bm{P_{\rm I}}$ & $P_{\rm MAX}$ & $P_{\rm OT}$ & $\bm{P_{\rm I}}$ & $P_{\rm MAX}$ & $P_{\rm OT}$ \\ \hline 
               $216.2$ & $206.1$ & $192.0$ & $280.1$ & $424.6$ & $430.3$ \\
	\end{tabular}
\end{table}

%%%%%%%%%%%%%%%%%%%%%%%%%%%%%%%%%%%%%%%%%%%%%%%%%%%%%%%%%%%%%%%%%%%%%%%%%%%%%
\section{Conclusion}
\label{sec:concl}
In this study, we theoretically discussed the following: (1) the extension of OT to I-OT, and (2) the cost robustness obtained using the I-OT solution. 
Subsequently, based on applications to logistics, we confirmed that the mathematical incorporation of prior knowledge enhances the robustness of transport.

The investigation performed in this work serves as a foundation for further research. 
A mismatch exists between the demand and available supply, which can be characterized by \emph{unbalanced} OT \cite{peyre2019computational}. 
Moreover, implicitly inferring the uncertainties using the currently adopted transportation plan is important. 
This is an \emph{inverse problem} that determines $Q$ when the optimal solution $P^*$ to Problem \ref{prob:robust} with (\ref{eq:robust_set}) and a given $C$ is available.

%%%%%%%%%%%%%%%%%%%%%%%%%%%%%%%%%%%%%%%%%%%%%%%%%%%%%%%%
%\clearpage

\bibliographystyle{IEEEtran}
\bibliography{IEEE_IOT}

\vfill

\end{document}